\documentclass[conference]{IEEEtran}

\IEEEoverridecommandlockouts

\usepackage{cite}
\usepackage{amsmath,amssymb,amsfonts}
\usepackage{algorithmic}
\usepackage{graphicx}
\usepackage{textcomp}
\usepackage{xcolor}
\usepackage{booktabs}
\usepackage{multirow}
\def\BibTeX{{\rm B\kern-.05em{\sc i\kern-.025em b}\kern-.08em
    T\kern-.1667em\lower.7ex\hbox{E}\kern-.125emX}}
\begin{document}

\title{Real-Time Underwater Image Enhancement via Frequency-Guided Dual-Path Attention}

\author{
\IEEEauthorblockN{
Leshen Zhang\textsuperscript{1,3}\IEEEauthorrefmark{1},
Ao Li\textsuperscript{2}\IEEEauthorrefmark{1},
Ce Zhu\textsuperscript{1,2}\IEEEauthorrefmark{2}
}
\IEEEauthorblockA{\textsuperscript{1}Glasgow College, UESTC, Chengdu, China}
\IEEEauthorblockA{\textsuperscript{2}School of Information and Communication Engineering, UESTC, Chengdu, China}
\IEEEauthorblockA{\textsuperscript{3}University of Glasgow, Glasgow, UK}

\IEEEauthorblockA{Email: lethyzhang@163.com, lethyacademic@gmail.com; aoli@std.uestc.edu.cn; eczhu@uestc.edu.cn}
\thanks{\IEEEauthorrefmark{1}Equal contribution. \IEEEauthorrefmark{2}Corresponding author.}
}

\maketitle

\begin{abstract}
Real-time underwater image enhancement (UIE) is crucial for mobile underwater photography and autonomous robotic systems, where practical deployment typically requires low latency and compact models under constrained computational resources. Recent ultra-lightweight CNNs based on structural re-parameterization meet these constraints but operate purely in the spatial domain, ignoring the frequency-sensitive nature of underwater degradation. To address this, we propose a lightweight UIE framework that integrates two key components: a {Multi-Branch Reparameterizable Convolution with Fixed DCT Priors (MBRConv-DCT)} that injects structured directional frequency priors during training, and a {Frequency-Guided Dual-Path Attention (FGDPA)} module that fuses spatial and spectral cues via a dual-path design for adaptive feature modulation.  Both components are fully compatible with structural re-parameterization: the convolution branch introduces \textbf{zero} additional inference cost after re-parameterization, while the attention module incurs only a minimal computational overhead.  Experiments show our model achieves state-of-the-art performance with only 4.23K parameters and 600+ FPS, outperforming much larger methods in both quantitative metrics and visual quality. Code is available at https://github.com/LethyZhang/FGDPA.
\end{abstract}


\section{Introduction}
\label{sec:intro}
Real-time underwater image enhancement (UIE) is critical for applications such as mobile underwater photography, autonomous underwater vehicles (AUVs), and embedded vision systems, where models must operate under strict constraints on latency, memory, and power~\cite{yan2026revisiting,bai2025towards}. Due to wavelength-dependent light absorption and scattering, underwater images suffer from severe degradation, including blurred details, color casts, and low contrast, which significantly impair downstream tasks like object detection and navigation~\cite{yan2025zero, yan2025plastic, yan2026animeagent}.

\begin{figure}
    \centering
    \includegraphics[width=0.75\linewidth]{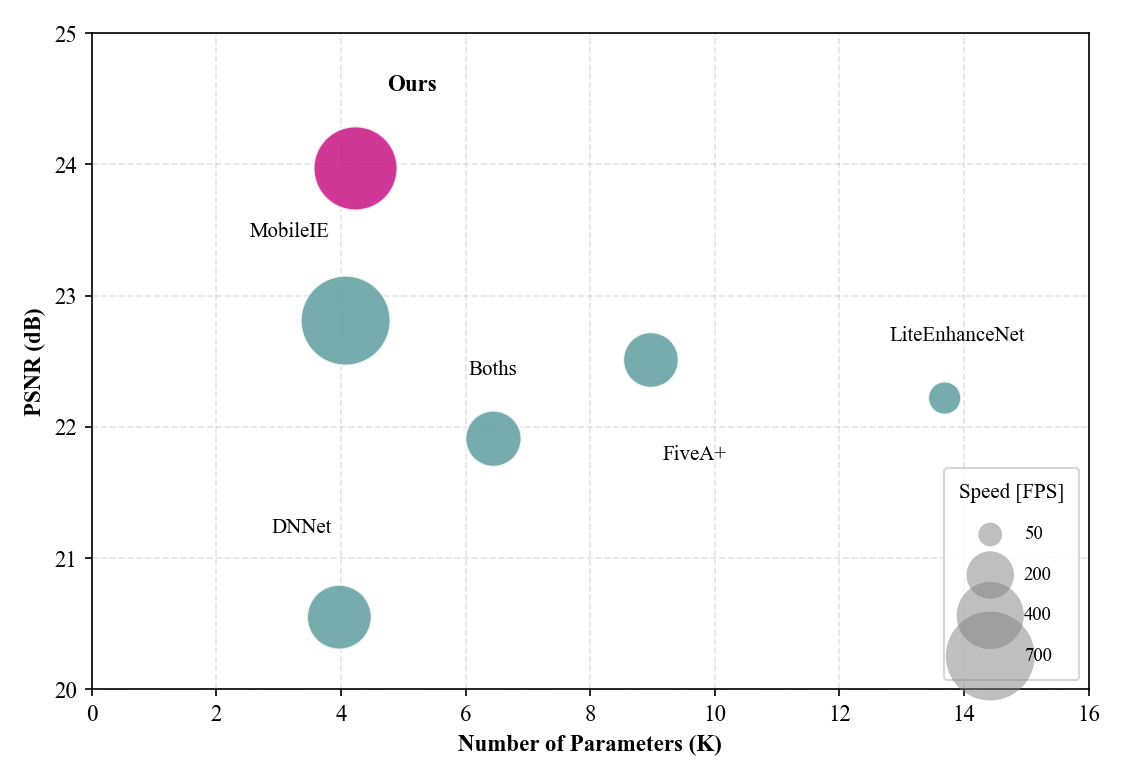}
    \caption{Comprehensive efficiency-effectiveness trade-off comparison among state-of-the-art image restoration methods.}
    \vspace{-4mm}
    \label{fig:placeholder}
\end{figure}

To address these challenges, early approaches primarily relied on handcrafted physical priors and contrast-based heuristics~\cite{shaahid2025aquadiffdiffusionbasedunderwaterimage}. While interpretable, their performance is often limited by inaccurate assumptions. Recent deep learning advances have surpassed these methods, with Transformer-based models achieving impressive visual quality by exploiting long-range dependencies and generative priors~\cite{liu2024ntire, liu2025ntire2025challenge}. However, these heavy architectures involve massive computation and parameter counts, rendering them impractical for resource-constrained edge deployment. Consequently, ultra-lightweight Convolutional Neural Networks (CNNs) based on structural re-parameterization—such as RepVGG~\cite{ding2021repvgg}, MobileOne~\cite{vasu2023mobileone}, and MobileIE~\cite{Yan_2025_ICCV}—have emerged as promising solutions. By collapsing multi-branch structures into single convolutions during inference, these models achieve millisecond-level latency with only a few thousand parameters. As illustrated in Fig.~\ref{fig:placeholder}, the efficiency–effectiveness map reveals a clear gap: while existing ultra-lightweight re-parameterized models deliver excellent speed and compactness, their restoration quality still plateaus below the desirable level, indicating that simply improving efficiency is insufficient without explicitly modeling frequency information.

Despite their efficiency, existing re-parameterized models often produce suboptimal results in complex underwater scenes, exhibiting washed-out colors and incomplete texture recovery~\cite{cong2024comprehensive}. We identify the core limitation as the absence of explicit frequency-domain awareness. Underwater degradation is inherently frequency-sensitive: absorption reshapes the global spectral distribution (affecting color), while scattering attenuates high-frequency components (eroding edges)~\cite{Zhao_2024_CVPR}. While frequency-domain representations (e.g., FFT, DCT) effectively separate these components for enhancement~\cite{11251228}, most existing frequency-aware methods are designed for large-scale networks, introducing computational overhead incompatible with real-time systems~\cite{s25103085}. In particular, frequency-aware attention frameworks such as FcaNet\cite{Qin_2021_ICCV} typically contain tens of millions of parameters, making them impractical for ultra-lightweight ($<10$K parameters), real-time models. Meanwhile, classical lightweight attentions such as SENet\cite{hu2018squeeze} and CBAM\cite{Woo_2018_ECCV} operate purely in the channel domains without explicit frequency modeling, and therefore are limited in addressing frequency-sensitive underwater degradation. Conversely, current lightweight UIE models operate purely in the spatial domain using isotropic convolutions~\cite{zhao2024toward}. Furthermore, the re-parameterization process—where branches merge into a single kernel—tends to diminish the directional and spectral diversity required to distinguish specific degradation modes. Since increasing model depth is infeasible under deployment constraints~\cite{BAI2026129125}, performance gains must stem from improving the representational efficiency of each parameter.

To overcome these limitations, we propose a frequency-guided lightweight UIE framework that integrates two key strategies: (1) injecting stable, structured spectral priors during training, and (2) leveraging global spectral statistics for lightweight feature modulation. Our approach introduces two complementary components: Multi-Branch Reparameterizable Convolution with Fixed DCT Priors (MBRConv-DCT), which enriches convolutions with directional frequency bases without inference overhead, and a Frequency-Guided Dual-Path Attention (FGDPA) module that performs adaptive, frequency-aware feature enhancement using compact FFT-based statistics.
The main contributions of this work are summarized as follows:
\begin{itemize}
\item We propose \textbf{Frequency-Guided Dual-Path Attention (FGDPA)}, an ultra-lightweight attention module that constructs a compact, resolution-invariant spectral descriptor from fixed-resolution FFT magnitude. It leverages global spectral statistics to guide channel modulation directly in the frequency domain, avoiding the heavy computational cost of inverse FFTs while capturing global degradation cues.
\item We introduce \textbf{MBRConv-DCT}, a multi-branch re-parameterizable convolution that injects fixed non-DC $3\times3$ DCT directional priors during training. This linear branch is analytically collapsible into standard $3\times3$ kernels at inference, effectively enriching the spectral diversity of the model without adding any parameters or FLOPs during deployment.
\item We demonstrate that explicit frequency guidance can be seamlessly integrated into ultra-lightweight, re-parameterized CNNs. Our model achieves state-of-the-art performance on UIEB with only 4.23K parameters and 600+ FPS, outperforming much larger models in both quantitative metrics and visual quality.
\end{itemize}

\section{Method}

\begin{figure}[!t]
    \centering
    \includegraphics[width=.45\textwidth]{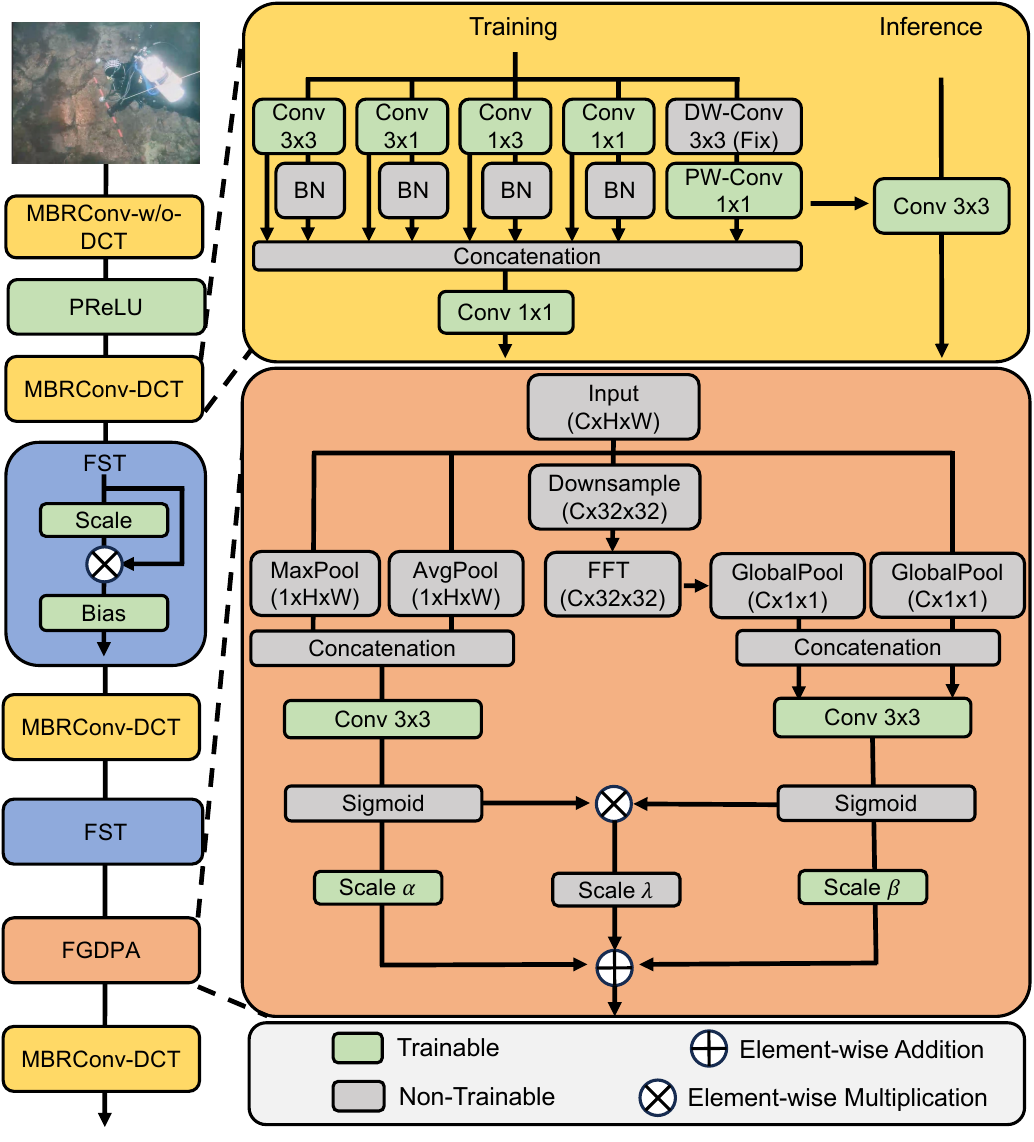}
    \caption{Overall architecture of the proposed framework. The network adopts a re-parameterizable training structure and collapses into a lightweight feed-forward model during inference.}
    \label{fig:overview}
\end{figure}

\subsection{Overview}

We present a frequency-guided ultra-lightweight UIE framework that enhances both frequency representation and frequency-aware modulation while preserving full compatibility with structural re-parameterization. (1) \textbf{MBRConv-DCT}, a re-parameterizable convolution module that injects fixed non-DC $3\times3$ DCT directional priors during training and fully collapses into standard $3\times3$ convolution kernels at inference; and (2) \textbf{Frequency-Guided Dual-Path Attention (FGDPA)}, which explicitly constructs a compact, resolution-invariant spectral magnitude representation on a deterministic $32\times32$ grid and uses it as global frequency guidance to modulate channel responses, while a parallel spatial branch focuses on saliency modeling in the spatial domain. Following MobileIE~\cite{Yan_2025_ICCV}, we retain the original Feature Self-Transform (FST) module, which enhances nonlinear feature interactions via a simple quadratic self-transformation. Both components add negligible training overhead, while the convolution branch incurs zero inference cost and the attention module introduces only minimal runtime overhead. 

\subsection{Multi-Branch Reparameterizable Convolution with Fixed DCT Priors}
To enhance the spectral representational capacity of re-parameterized convolutions, we propose {MBRConv-DCT}, an extension of the Multi-Branch Reparameterizable Convolution (MBRConv) that injects fixed directional frequency priors through a dedicated DCT branch. 

The core idea is to complement learnable spatial filters with a set of precomputed, non-DC $3\times3$ Discrete Cosine Transform (DCT) basis kernels. These kernels encode structured directional responses—including horizontal, vertical, and diagonal patterns—spanning low to high spatial frequencies, while excluding the DC (mean) component to avoid redundancy with the main convolutional branches. Each kernel is normalized to zero mean and unit $\ell^2$ norm, ensuring numerical stability and preserving feature statistics.

As illustrated in Figure~\ref{fig:overview}, MBRConv-DCT consists of the original MBRConv parallel branches (i.e., $3\times3$, $3\times1$, $1\times3$ and $1\times1$ convolutions, each followed by batch normalization) and an additional \textit{linear DCT branch}. Given an input feature map $F$, this branch computes:
\begin{equation}
F_{\text{DCT}} = \text{DWConv}_{\text{DCT}}(F),
\end{equation}
where $\text{DWConv}_{\text{DCT}}$ applies depthwise convolution using the fixed DCT kernels (not trainable). The responses are then projected to the target channel space via a learnable pointwise convolution and modulated by a scalar gate:
\begin{equation}
F_{\text{proj}} = \tanh(\gamma) \cdot \text{PWConv}(F_{\text{DCT}}),
\end{equation}
with $\gamma$ initialized to zero. This ``zero-impact'' initialization ensures the branch does not disrupt training at the beginning, and is gradually activated only if beneficial.

Crucially, the DCT branch is entirely linear and can be analytically fused into an equivalent $3\times3$ convolution kernel during re-parameterization. Specifically, the composition of the fixed depthwise kernels and the learned pointwise weights yields a standard $3\times3$ kernel per output channel. This equivalent kernel is concatenated with the re-parameterized kernels from all other parallel branches and merged through the final $1\times1$ fusion layer. As a result, MBRConv-DCT introduces {no additional parameters or FLOPs} at inference time, while enriching the model’s sensitivity to directional frequency cues during training.

\subsection{Frequency-Guided Dual-Path Attention}
To inject resolution-invariant spectral priors into attention generation under strict efficiency constraints, we propose a Frequency-Guided Dual-Path Attention (FGDPA) module. Given an intermediate feature map $F \in \mathbb{R}^{B \times C \times H \times W}$, FGDPA first constructs a deterministic low-resolution frequency representation to ensure stability across different input resolutions. Specifically, $F$ is downsampled to a fixed $32 \times 32$ grid using non-adaptive average pooling, and a 2D FFT is applied to obtain the log-compressed magnitude spectrum:
\begin{equation}
M = \log(1 + |\mathcal{F}(F_{32\times 32})|),
\end{equation}
where the logarithmic transform stabilizes the dynamic range and the fixed resolution eliminates spectral variance caused by size changes.
Unlike complex frequency-band decomposition strategies, we do not explicitly separate or reconstruct different frequencies. Instead, we directly aggregate the frequency magnitude to capture global spectral statistics. While spatial pooling focuses on intensity, pooling the logarithmic magnitude spectrum $M$ allows the network to perceive the overall spectral energy distribution. Global average pooling is applied to $M$, followed by normalization to suppress scale bias, and concatenated with the global average pooled spatial feature $\text{GAP}(F)$. A lightweight $1\times1$ convolution with sigmoid activation then yields a frequency-guided channel attention map $A_c \in \mathbb{R}^{B \times C \times 1 \times 1}$, enabling channel responses to be adaptively modulated according to the global spectral energy distribution.
In parallel, a spatial attention pathway is constructed purely in the spatial domain to emphasize salient structural regions. The channel-wise maximum and average projections of $F$ are concatenated and fed into a $1\times1$ convolution followed by a sigmoid function to produce a spatial attention map $A_s \in \mathbb{R}^{B \times C \times H \times W}$.
These two attention branches constitute the ``dual paths'' of FGDPA: a \emph{frequency-guided channel path} and a \emph{spatial saliency path}. Instead of relying on dynamic routing, we employ a static, learnable hybrid gating to combine them:
\begin{equation}
A = (1 - \lambda)(\alpha A_c + \beta A_s) + \lambda (A_c \odot A_s),
\end{equation}
Here, $A_c \in \mathbb{R}^{B\times C\times 1\times 1}$ and $A_s \in \mathbb{R}^{B\times 1\times H\times W}$. The summation term $(\alpha A_c + \beta A_s)$ adopts standard broadcasting, where $A_c$ is spatially expanded and $A_s$ is channel-wise replicated. The multiplicative term $(A_c \odot A_s)$ follows the same broadcasting semantics, resulting in the final modulation tensor $A \in \mathbb{R}^{B\times C\times H\times W}$ where $\alpha, \beta \in \mathbb{R}$ and $\lambda \in [0,1]$ are learnable scalars. The final modulated feature is obtained as $F' = A \odot F$.
\begin{table*}[!t]
\centering
\setlength{\tabcolsep}{5pt}
\renewcommand{\arraystretch}{1.2}
\caption{Comprehensive comparison under 480$\times$640 resolution. Best and second best are marked in \textcolor{red}{red} and \textcolor{blue}{blue}, respectively.}
\label{tab:final_compare}
\resizebox{\textwidth}{!}{
\begin{tabular}{l c c c c | ccccc | cc | cc}
\toprule
\multirow{2}{*}{Method} &
\multirow{2}{*}{\#Params (K)$\downarrow$} &
\multirow{2}{*}{FLOPs (G)$\downarrow$} &
\multirow{2}{*}{FPS$\uparrow$} &
\multirow{2}{*}{Venue} &
\multicolumn{5}{c|}{\textbf{UIEB (Supervised)}} &
\multicolumn{2}{c|}{\textbf{EUVP (Unsupervised)}} &
\multicolumn{2}{c}{\textbf{UIQS (Unsupervised)}} \\
\cmidrule(lr){6-10} \cmidrule(lr){11-12} \cmidrule(lr){13-14}
& & & & &
PSNR$\uparrow$ & SSIM$\uparrow$ & LPIPS$\downarrow$ & UIQM$\uparrow$ & UCIQE$\uparrow$
& UIQM$\uparrow$ & UCIQE$\uparrow$
& UIQM$\uparrow$ & UCIQE$\uparrow$ \\
\midrule

Boths\cite{9991160}& 6.447 & 1.9738 & 281.72 & GRSL'23 
& 21.9103 & 0.8981 & 0.1519 & \textcolor{blue}{1.8790} & 0.3418
& 1.7305 & \textcolor{blue}{0.3220} 
& 1.4923 & \textcolor{red}{0.3252} \\

U-Shape\cite{10129222}& 31589.7 & 13.9164 & 11.19 & TIP'23
& 22.5180 & 0.8229 & 0.1791 & 1.7989 & \textcolor{blue}{0.3476}
& 1.6799 & 0.3076
& 1.3899 & 0.2943 \\

LiteEnhanceNet\cite{ZHANG2024122546}& 13.688 & 3.2342 & 98.59 & ESWA'24
& 22.2205 & \textcolor{blue}{0.9142} & \textcolor{blue}{0.1318} & 1.8712 & 0.3312
& \textcolor{blue}{1.7622} & 0.3013
& 1.4803 & 0.2922 \\

LSNet~\cite{zhou20247k}
& 8.245 & 10.2995 & 9.64 & arXiv'24
& 18.0125 & 0.8204 & 0.2413 & 1.7807 & 0.2511
& 1.6872 & 0.2840
& \textcolor{red}{1.5830} & 0.2827 \\

MobileIE~\cite{Yan_2025_ICCV}
& 4.075 & 1.1833 & {718.62} & ICCV'25
& \textcolor{blue}{22.8050} & 0.9098 & 0.1414 & 1.8695 & 0.3443
& 1.7451 & 0.3128 
& 1.5069 & 0.2838 \\

{Ours} 
& 4.234& 1.1796& {628.25}& --
& \textcolor{red}{23.9660} & \textcolor{red}{0.9155} & \textcolor{red}{0.1188} & \textcolor{red}{1.9051} & \textcolor{red}{0.3525}
& \textcolor{red}{1.7954} & \textcolor{red}{0.3257}
& \textcolor{blue}{1.5220} & \textcolor{blue}{0.2956} \\
\bottomrule
\end{tabular}}
\end{table*}
\begin{figure*}
    \centering
    \includegraphics[width=0.9\linewidth]{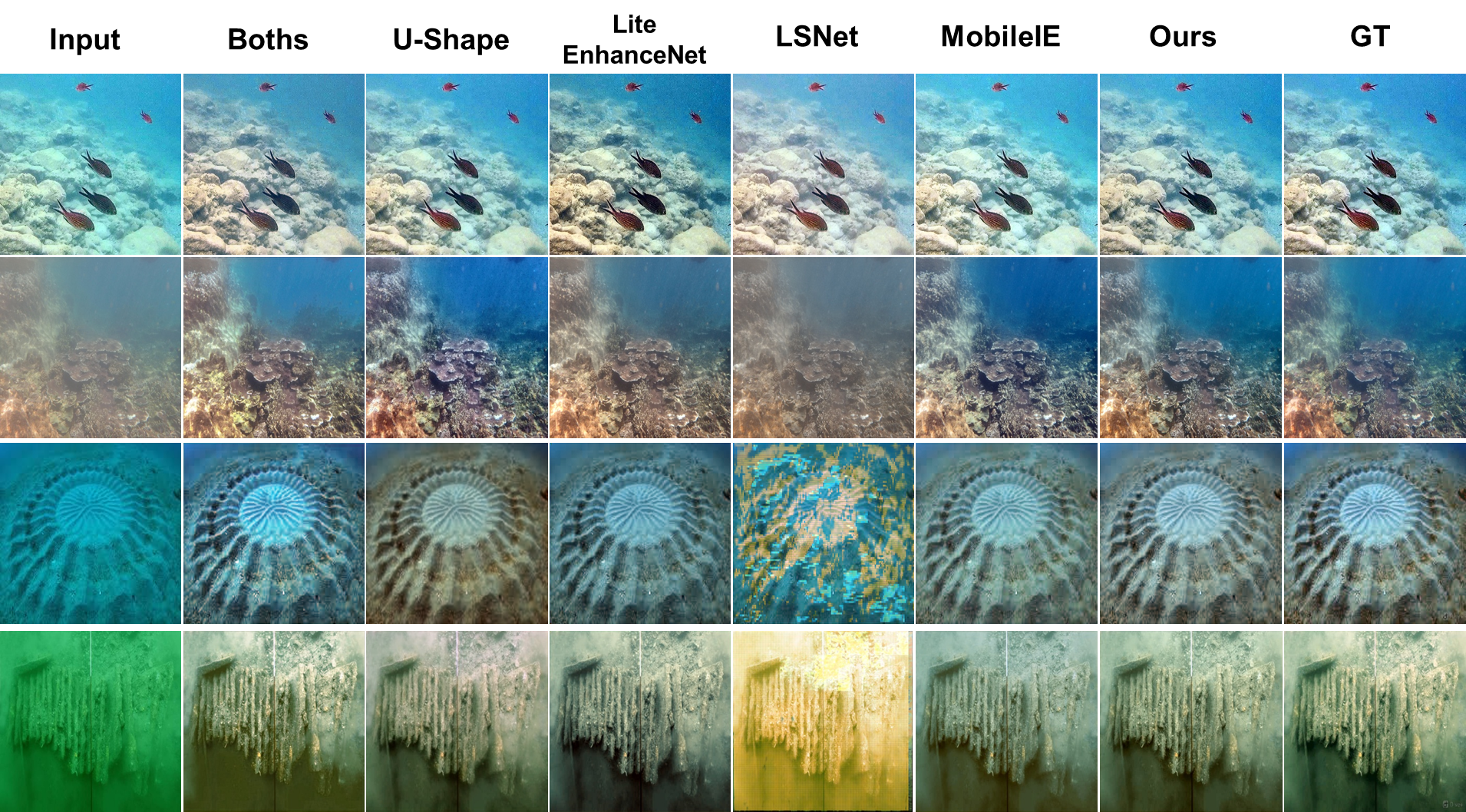}
    \caption{Qualitative comparison of underwater image enhancement results on the UIEB dataset with ground-truth references. }
    \label{fig:qualitative}
\end{figure*}
\begin{figure}
    \centering
    \includegraphics[width=0.75\linewidth]{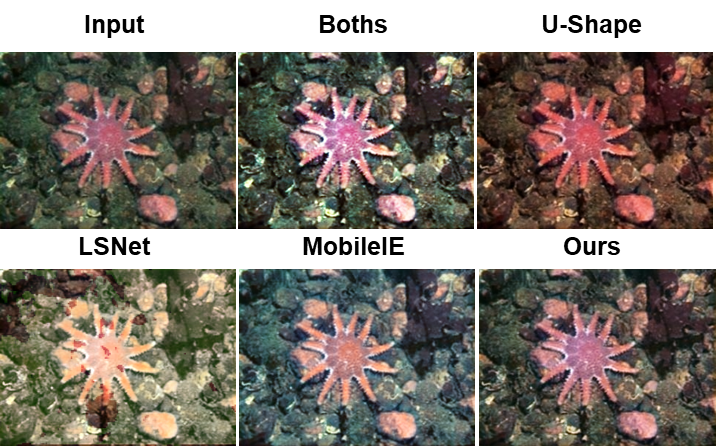}
    \caption{Qualitative comparison of underwater image enhancement results on real-world EUVP images without ground-truth references. }
    \label{fig:qualitative2}
\end{figure}
Importantly, the entire frequency branch operates only at $32\times32$ resolution and requires no inverse FFT. All convolutions are standard $1\times1$ layers, ensuring FGDPA introduces minimal overhead during training and minimal additional cost at inference.

\section{Experiments}
\subsection{Experimental Settings}
\noindent\textbf{Datasets.} All experiments are conducted on the UIEB dataset~\cite{8917818}, a widely used benchmark for underwater image enhancement that contains 890 paired underwater–reference image pairs. Following common practice, we use 800 images for training and 90 images for testing.  In addition to reference-based evaluation on UIEB, we further assess generalization performance on two non-reference real-world datasets: the EUVP dataset~\cite{islam2020fast}, which comprises 1,868 real underwater images captured under diverse turbidity and lighting conditions without ground truth, and the UIQS, a no-reference subset of UIEB used to evaluate enhancement quality in realistic scenarios where ground truth is unavailable.

\noindent\textbf{Implementation Details.} All models are implemented in PyTorch and trained on a single NVIDIA RTX 3090 GPU. We adopt the Adam optimizer with $\beta_1 = 0.9$ and $\beta_2 = 0.999$. The initial learning rate is set to $1\times10^{-3}$. During training, input images are randomly cropped into $256\times256$ patches, followed by random horizontal flipping and rotation for data augmentation. The batch size is set to 4, and all models are trained for 1,000 epochs. The proposed method uses a fixed channel width of $C = 12$. Its attention module (FGDPA) computes FFT guidance on a deterministic $32 \times 32$ downsampled feature map, with fusion weights initialized as $\alpha = 1.0$, $\beta = 1.0$, and $\lambda = 0.5$. All MBRConv-DCT layers use a branch expansion ratio of 4 and integrate $K = 4$ fixed, non-DC $3\times3$ DCT kernels. We adopt the same Local Variance-Weighted (LVW) loss as MobileIE~\cite{Yan_2025_ICCV}, ensuring reproducibility and fair comparison.

\subsection{Comparison with SOTA Methods}
We compare our method with representative state-of-the-art approaches, ranging from large-scale restoration networks (e.g., U-Shape~\cite{10129222}) to recent ultra-lightweight mobile designs (e.g., Boths~\cite{9991160}, LiteEnhanceNet~\cite{ZHANG2024122546}, LSNet~\cite{zhou20247k}, and MobileIE~\cite{Yan_2025_ICCV}).  All compared methods are evaluated under a unified and fair protocol. Quantitative results on the supervised UIEB benchmark (Table~\ref{tab:final_compare}) demonstrate that our model achieves the best overall performance, reaching 23.97 dB PSNR and 0.9155 SSIM. Notably, we outperform the strongest lightweight competitor, MobileIE, by over 1.1 dB while maintaining a comparable parameter count (~4.2K). Our method even surpasses the computation-heavy U-Shape Transformer (31.5M parameters) in both distortion-based and perceptual metrics (achieving the lowest LPIPS of 0.1188).

To assess generalization in real-world scenarios, we further evaluated our method on two non-reference datasets: EUVP and UIQS. As shown in Table~\ref{tab:final_compare}, our model demonstrates superior robustness in the wild. On the EUVP dataset, we achieve state-of-the-art performance with the highest UIQM (1.7954) and UCIQE (0.3257) scores. On the challenging UIQS dataset, our method consistently outperforms the baseline MobileIE and ranks second overall among all methods. This confirms that our frequency-guided design effectively handles diverse, heterogeneous underwater degradations beyond the training distribution.

\textbf{Qualitative Results.} Figure~\ref{fig:qualitative} presents visual comparisons on challenging UIEB test images with available ground-truth references. Compared with representative SOTA approaches, several lightweight CNN-based models still suffer from insufficient contrast enhancement and loss of structural details. In contrast, our method yields more visually faithful reconstructions, achieving better color fidelity, clearer object boundaries, and improved scene visibility, particularly in heavily degraded regions. These visual results align well with the quantitative evaluations on UIEB, further validating the effectiveness of the proposed FGDPA framework.

To further assess generalization under practical, unconstrained underwater conditions, Figure~\ref{fig:qualitative2} shows qualitative results on real-world EUVP images without ground-truth references. The scenes in EUVP exhibit severe color casts, turbidity, and heterogeneous degradations, where many competing methods suffer from residual haze, color imbalance, or unnatural visual artifacts. By contrast, our approach consistently produces visually pleasing restorations with balanced color tones, enhanced global contrast, and preserved texture details, demonstrating strong robustness and applicability in real underwater environments.

\begin{table}[!t]
\renewcommand{\arraystretch}{1.5}
\setlength{\tabcolsep}{8pt}
\caption{Ablation study on the UIEB dataset.}
\label{tab:ablation}
\centering
\resizebox{.9\columnwidth}{!}{
\begin{tabular}{l c c c}
\toprule
Method & PSNR$\uparrow$ & SSIM$\uparrow$ & LPIPS$\downarrow$ \\
\midrule
(a) Baseline & 18.96& 0.8649& 0.2294\\
(b) Baseline + MBRConv-DCT & 19.28 & 0.8660 & 0.2197 \\
(c) Baseline + FGDPA & 23.67 & 0.9121 & 0.1237 \\
(d) Ours - frequency magnitude & 22.81 & 0.9093 & 0.1355 \\
(e) Ours & \textbf{23.97}& \textbf{0.9155}& \textbf{0.1188}\\
\bottomrule
\end{tabular}
}
\end{table}
\begin{figure}[!t]
    \centering
    \includegraphics[width=\linewidth]{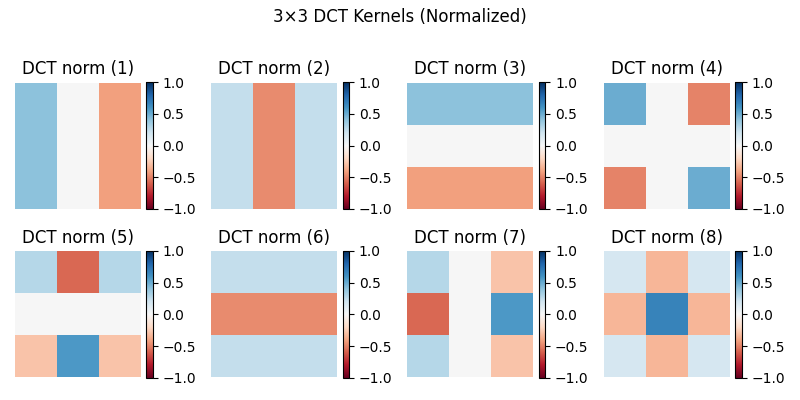}
    \caption{Visualization of the eight normalized $3\times3$ non-DC DCT kernels in MBRConv-DCT. Each kernel is zero-mean and unit-norm, encoding distinct directional frequency priors that complement learnable convolutions with fixed spectral inductive bias.}
    \label{fig:dct_kernels}
\end{figure}

\subsection{Ablation Study}
We conduct an ablation study on the UIEB dataset to validate the contribution of each component (Table~\ref{tab:ablation}). 
In the ablation study, the baseline refers to our backbone architecture built upon MobileIE, where proposed attention mechanisms are removed, keeping only the basic re-parameterizable convolution pipeline. Introducing the proposed \textit{MBRConv-DCT} (row b) improves PSNR by +0.32~dB over the baseline, confirming that fixed non-DC DCT kernels enrich spectral diversity. As visualized in Figure~\ref{fig:dct_kernels}, these kernels encode directional responses (horizontal, vertical, and diagonal), which are highly consistent with underwater degradation patterns, while introducing zero inference cost owing to the re-parameterization design. Note that the number of non-DC DCT kernels $K$ is configurable; we visualize eight kernels ($K=8$) in Figure~\ref{fig:dct_kernels} for clearer illustration, while $K=4$ is adopted in all experiments to strike a better accuracy–efficiency balance. We further observe that performance remains stable across $K=2$ to $K=8$ (within $\pm 0.1$ dB), indicating that the DCT branch provides a robust spectral inductive bias rather than a sensitive hyperparameter. Equipping the network with \textit{FGDPA} (row c) leads to a dramatic +4.71dB PSNR gain compared with the baseline, demonstrating that explicit frequency guidance is crucial: frequency guidance stabilizes global color restoration and improves structural fidelity by providing complementary spectral statistics that are unavailable to purely spatial attention. We also analyze the effect of FFT resolution: increasing the resolution beyond $32\times32$ brings only a marginal gain (+0.21 dB PSNR) with much higher computational cost, while reducing the resolution to $16\times16$ leads to a noticeable performance drop (-0.49 dB PSNR), validating the chosen $32\times32$ configuration as the optimal trade-off between performance and efficiency. To further examine the role of frequency priors, we remove the frequency magnitude from the channel attention pathway (row d). This degradation results in a noticeable performance drop (23.97~dB $\rightarrow$ 22.81~dB), indicating that frequency-aware channel modulation is indispensable rather than merely auxiliary. Finally, the full model (row e) achieves the best overall performance (23.97~dB PSNR and the lowest error), validating their complementary roles: MBRConv-DCT enhances representational richness, while FGDPA enables fine-grained, frequency-aware feature modulation.

\subsection{Complexity Analysis}
We analyze the efficiency of our framework both theoretically and empirically. Let the input feature map have dimensions $H\times W\times C$, and the backbone contain $N$ convolutional blocks. As in MobileIE, the dominant cost is the re-parameterized $3\times3$ convolutions, yielding $\mathcal{O}(NC^2HW)$ FLOPs. The {MBRConv-DCT} branch is fully linear and analytically merged into the $3\times3$ kernel during re-parameterization, introducing {zero additional parameters or FLOPs at inference}. The FGDPA module computes frequency guidance on a fixed $32\times32$ grid, so the frequency-analysis part has a constant complexity independent of input resolution, while the remaining attention operations scale linearly with $H$ and $W$. Its operations include: (1) deterministic downsample, (2) $32\times32$ FFT (complexity $\mathcal{O}(C \cdot 32^2 \log 32)$, a constant), and (3)two $1\times1$ convolutions ($2CHW + 2C^2$ FLOPs), plus $\mathcal{O}(CHW)$ statistics and element-wise operations. Thus, FGDPA adds only a {negligible constant-factor overhead}, preserving the asymptotic complexity $\mathcal{O}(NC^2HW)$. Empirically, our model has only {4.234K parameters}. At $256\times256$ and $480\times640$ resolutions, it achieves {0.2517} and {1.1796 GFLOPs}, with {1142.75} and {628.25 FPS} on an NVIDIA RTX 3090, confirming its ability to achieve real-time throughput under desktop GPU settings. To further validate practical deployment capability, we evaluate the proposed model on an embedded platform (Jetson AGX Orin 64GB). At resolution $640\times480$, the model achieves 106.9 FPS with a latency of 9.35 ms. These results demonstrate that the proposed design not only achieves theoretical efficiency but also sustains real-time performance under edge constraints, as summarized in Table~\ref{tab:edge_deployment}.

\begin{table}[h]
\centering
\caption{Jetson AGX Orin 64GB deployment.}
\label{tab:edge_deployment}
\begin{tabular}{c c c c c}
\toprule
Res. & FPS & Latency & Params & GFLOPs \\
\midrule
$640{\times}480$ & \textbf{106.9} & \textbf{9.35 ms} & 4.23K & 1.18 \\
\bottomrule
\end{tabular}
\vspace{-4mm}
\end{table}

\section{Conclusion}
In this paper, we have demonstrated that frequency modeling can be effectively integrated into ultra-lightweight underwater image enhancement without sacrificing real-time performance. By incorporating explicit frequency guidance through structured spectral priors and lightweight FFT-based frequency statistics, our method effectively strengthens frequency awareness in ultra-lightweight CNNs without sacrificing real-time performance.  With only {4.234K parameters} and {600+ FPS} at $480\times640$ resolution, it achieves consistent gains over existing mobile-friendly approaches, demonstrating strong potential for deployment on resource-constrained platforms such as underwater drones and embedded robotic systems.

\section*{Acknowledgment}

This research was supported by the Hainan Provincial Joint Project of Li'an International Education Innovation Pilot Zone under Grant No. 624LALH003, and the Student Science and Technology Innovation Fund of the Glasgow College, UESTC.

\bibliographystyle{IEEEtran}
\bibliography{IEEEabrv}

\end{document}